\documentclass[a4paper, 12pt, titlepage=false]{article}
\usepackage[english]{babel}
\usepackage[utf8]{inputenc}
\usepackage[T1]{fontenc}
\usepackage{setspace}
\usepackage{datetime}
\usepackage{mathptmx}
\usepackage{graphicx}
\usepackage[utf8]{inputenc}
\usepackage[top=2.5cm, left=2.5cm, bottom=2.5cm, right=2.5cm]{geometry} 
\usepackage{amsmath,amsthm,amssymb}
\usepackage{url} 
\usepackage{csquotes}
\usepackage{natbib}
\usepackage{indentfirst}
\usepackage[colorlinks = true, linkcolor = black, citecolor=black, urlcolor=black, filecolor = black, bookmarks=false,hypertexnames=false]{hyperref}
\usepackage[none]{hyphenat} 
\bibpunct{\textcolor{black}{[}}{\textcolor{black}{]}}{;}{a}{}{,}
\usepackage{float}
\usepackage{comment}
\usepackage{colortbl}
\usepackage{subcaption} 
\usepackage{caption} 
\usepackage{titling} 
\usepackage{hyperref}

\usepackage{times}
\newcommand{\YourTitleReport}{Peer-Ranked Precision: Creating a Foundational Dataset for Fine-Tuning Vision Models from DataSeeds’ Annotated Imagery}

\begin{document}

\begin{center}
\large\textbf{\YourTitleReport}

\vspace{0.5cm}
\normalsize
\textbf{Sajjad Abdoli\textsuperscript{1*}, 
Freeman Lewin\textsuperscript{2}, Gediminas Vasiliauskas\textsuperscript{3}, Fabian Schonholz\textsuperscript{4}}

\vspace{0.2cm}
\small
\textsuperscript{1}Perle.ai, \href{mailto:sajjad@perle.ai}{sajjad@perle.ai}\\
\textsuperscript{2}Emet Research, \href{mailto:freeman@emetresearch.ai}{freeman@emetresearch.ai} \\
\textsuperscript{3}Zedge, \href{mailto:gediminas.vasiliauskas@zedge.net}{gediminas.vasiliauskas@zedge.net}\\
\textsuperscript{4}FESSEX, \href{mailto:fabian@fessexconsulting.com}{fabian@fessexconsulting.com} \\
\textsuperscript{*}Corresponding author
\end{center}

\pagenumbering{arabic}
\setcounter{page}{1}

\begin{center}
    \section*{Abstract}
    \end{center}
    \begingroup
    \small 
    \setlength{\parindent}{0pt} 
    \sloppy
    The development of modern Artificial Intelligence (AI) models, particularly diffusion-based models employed in computer vision and image generation tasks, is undergoing a paradigmatic shift in development methodologies. Traditionally dominated by a “Model Centric” approach, in which performance gains were primarily pursued through increasingly complex model architectures and hyperparameter optimization, the field is now recognizing a more nuanced “Data-Centric” approach.  This emergent framework foregrounds the quality, structure, and relevance of training data as the principal driver of model performance.  To operationalize this paradigm shift, we introduce the DataSeeds.AI sample dataset (the “DSD”), initially comprised of approximately 10,610 high-quality human peer-ranked photography images accompanied by extensive multi-tier annotations.  The DSD is a foundational computer vision dataset designed to usher in a new standard for commercial image datasets.  Representing a small fraction of DataSeeds.AI's 100 million-plus image catalog, the DSD provides a scalable foundation necessary for robust commercial and multimodal AI development.  Through this in-depth exploratory analysis, we document the quantitative improvements generated by the DSD on specific models against known benchmarks and make the code and the trained models used in our evaluation publicly available.\footnote{Experiments undertaken for this paper were done on the DSD which contained 10,610 images. Prior to publication, in order to account for a wide range of multi-jurisdictional compliance considerations, we removed 2,838 images that contained subjects of a sensitive nature, including people's faces. The remaining 7,772 images are available for AI training at \url{https://huggingface.co/datasets/Dataseeds/DataSeeds.AI-Sample-Dataset-DSD}}\footnote{We wish to thank DataSeeds.AI for their immense contribution to this project, which made it possible. As part of this project, a sister dataset of 10,000 fully annotated and segmented images is now immediately available for purchase.  For more information on this study, sales, or annotation, please contact \href{mailto:sales@dataseeds.ai}{sales@dataseeds.ai}.}
    \endgroup

    \section{Introduction}
    
    Recent scholarship has challenged long-standing assumptions within the AI community about where the most meaningful performance gains originate by demonstrating that a 'Data Centric' or 'Data Oriented' approach to model development can produce superior performance and, by extension, cost savings, across a wide range of tasks.  As articulated by ~\cite{bhatt2024data}, the Data Centric approach shifts the focus of model improvement from algorithmic refinement to deliberate curation and enhancements of training datasets.  In contrast to the traditional Model Centric approach, where researchers and developers concentrate on optimizing model architectures and tuning hyperparameters while treating the underlying training data as static, the Data Centric approach posits that systematic improvements to data quality, structure, and fidelity can produce more robust and generalizable models. 
    
    Despite research showing significant gains in model performance as a result of leveraging a Data Centric approach, practical implementation of data-centric strategies remains constrained by high operational costs, the difficulty of sourcing high-quality datasets at scale, and the persistent presence of noisy or duplicative data in widely used benchmarks. As highlighted in ~\cite{bhatt2024data}, constructing datasets with meaningful annotations and verified integrity often requires substantial manual effort and domain-specific knowledge. 
    
    The Dataseeds Sample Dataset (DSD) addresses these challenges head on.  Comprised of approximately 10,610 human peer-ranked photography images with multi-tier human annotation, the DSD and its broader commercially available datasets offer a rare confluence of both quality and quantity.  Human peer-based ranking introduces a novel, organically generated signal of perceptual and aesthetic value, while expert-guided annotation ensures that each image contributes reliable information across a vast range of computer vision tasks.  In contrast to widely used image datasets like ImageNet (\cite{deng2009imagenet}) or OpenImages (\cite{kuznetsova2020open}), which often rely on noisy crowd-sourced labels, the DSD is explicitly curated to minimize label noise, improve representational diversity, and maximize utility in multimodal and commercial model contexts.

    \section{DSD Datasource, Human Ranking, \& EXIF Metadata}
    
    The DSD is derived from GuruShots\footnote{https://gurushots.com/}, a popular online photography game that combines elements of social engagement, competition, and gamification.  The GuruShots platform hosts regular, themed photography challenges where photographers from diverse backgrounds, both amateur and professional, submit their photos for human peer review, conducted through a gamified voting experience which lends itself to creating a uniquely valuable image dataset reflecting human aesthetic preferences and technical evaluation, both valuable markers for reinforcement learning and other AI modeling approaches.

\begin{figure}[H]
        \centering
        \includegraphics[width=0.7\linewidth]{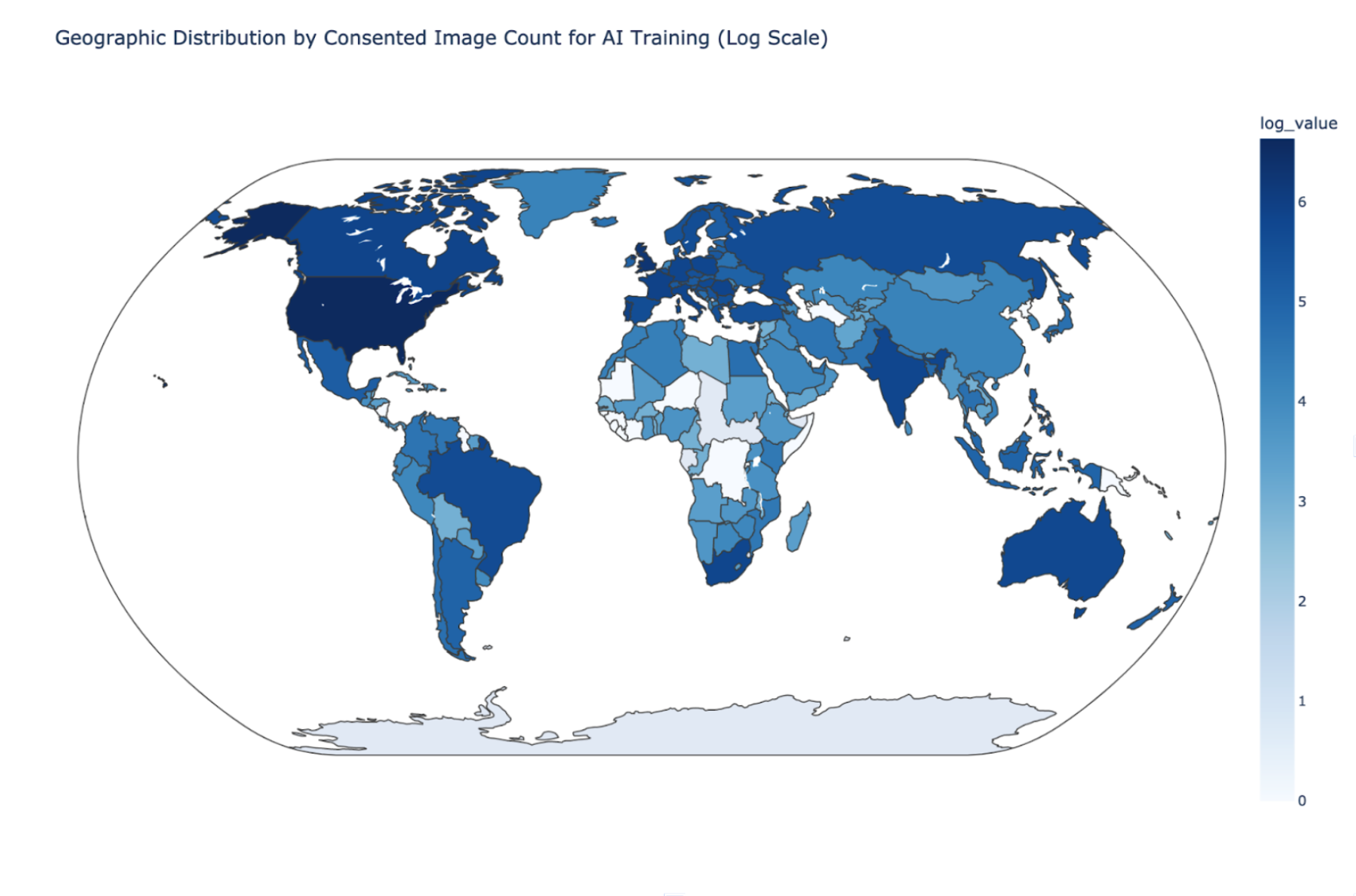}
        \caption[Geographic Distribution by Consented Image Count for AI Training]{Geographic Distribution by Consented Image Count for AI Training (Log Scale)}
        \label{fig:geo_dist}
    \end{figure}

    Critically, the DataSeeds.AI platform has the capacity to deliver fully-licensed on-demand image, video, and voice datasets to each client's spec. While the DSD was sourced from the GuruShots catalog of 100 million-plus high-quality images, DataSeeds.AI more generally, has the ability to source fully licensed photos, on demand, across a wide variety of categories and geographies. By default, each photo is accompanied by extensive EXIF data, including camera make, model, f\_number, aspect ratios, lens type, flash, focal length, and some cases, GPS coordinates. Recent testing of the GuruShots catalog revealed several unique attributes including images from 628 unique makes (Canon, Nikon, Apple, etc.), 8000+ models (Canon EOS 6D, NIKON D750, iPhone X, etc.), 9357 lenses  (iPhone back camera 4.15mm f/2.2, Canon EF 24-105mm f/4L IS, etc.), and a wide representation of common aspect ratios, such as 1:1, 16:9, 3:2, 4:3, and their inverse orientations, making it one of the most diverse commercially available datasets on the market. Figure \ref{fig:geo_dist} sets forth the geographic distribution of its consented image catalog, with United States users making up 17.98\% of consented photos, followed by the UK, Portugal, France, and lastly Slovenia (0.75\%). 
    
    The DSD uniquely integrates professional photographic standards with semi-professional creative expression, presenting a wide-ranging visual dataset with inherent hierarchical quality signals. This wealth of technical metadata in itself provides valuable context for AI models and facilitates in-depth analysis and comparison across diverse photographic techniques and conditions. Unlike many image data sources, the DSD (and the broader GuruShots dataset), when viewed in conjunction with its metadata, can be effective on its own, without explicit annotation.  Leveraging its built-in community-driven ranking system, each image inherently carries an evaluative dimension based on human preferences and perceptions.  Thus, beyond conventional image datasets, the DSD enables immediate usability for various tasks, including aesthetic evaluation, quality assessment, and stylistic categorization, without additional annotation efforts.
    
    \section{DSD Segmentation \& Annotation}
    
    Research has shown that human annotation can significantly reduce training costs for computer vision and multimodal artificial generative AI models, while simultaneously improving model accuracy and data efficiency (\cite{zahn2023efficient}). Recognizing this research and anticipating the growing adoption of human-in-the-loop annotation processes as standard within Data Centric organizations, we employed a complementary multi-tier human-in-the-loop annotation strategy, which closely resembles, but was independently developed prior to the publication and our awareness of, NVIDIA’s Describe Anything Model (DAM) framework (\cite{lian2025describe}) and its semi-supervised data pipeline (DLC-SDP).  Our approach, like the DLC-SDP, combined structured human annotations with scalable machine-assisted processes to enhance data quality and efficiency.
    Each image in the DSD underwent a meticulous multi-tier annotation process comprised of three tiers of distinct yet complementary human-crafted textual annotations and detailed semantic segmentation masks.
    
    \subsection{Semantic Segmentation}
    
    Whereas many commercially available image datasets restrict themselves to bounding box-level annotation, we choose to adopt a more rigorous standard: full semantic segmentation. Bounding boxes, while expedient, delineate crude rectangular regions and are prone to capturing irrelevant background artifacts.  In contrast, semantic segmentation defines the precise pixel-level contours of objects, capturing their true morphology, including the many occluded and overlapping forms inherent to the high-quality photography found throughout the DSD. 
    
    This type of granularity was chosen not merely for its aesthetic. It yielded more accurate scene understanding and enabled finer-grained feature extraction, both of which are critical to many of the anticipated use cases for the DSD that demand a high degree of spatial discernment.  By distinguishing objects from background at a fine resolution, segmentation enables models to infer depth, hierarchy, and adjacency relationships otherwise flattened by bounding box annotation.  As such, the segmentation masks found in the DSD, whose subject matter frequently includes compositional layering and nuanced object interrelations, renders the data particularly amenable to applications in generative image modeling, scene synthesis, and augmented reality, where the integrity of spatial relationships must be preserved.
    
    Our leveraging of segmentation masks within the annotation pipeline allowed us to enforce strict quality control protocols.  Each mask was subject to manual review and verification by trained annotators, guided by a checklist calibrated to minimize label noise and semantic ambiguity. This procedural rigor ensures the dataset’s reliability, allowing for consistent benchmarking and reproducible improvements in model training. While semantic segmentation allowed for technical enhancement, it also ensured high-fidelity, precision, and interpretability.
    
    \subsection{Annotation}
    The textual component of the annotation was structured into three distinct layers of human-crafted descriptions, designed with the anticipation that some users may simply want to capture the textual descriptions as their own dataset, separate from the images.  At the initial tier, concise descriptive English language titles were created to encapsulate the central subjects of each image, while being optimized for rapid reference and retrieval. The second tier provided for detailed narratives of the images of no fewer than fifteen words, systematically capturing object identifiers, spatial relationships, and an objective documentation of visual elements.  
    The final tier offered technical scene analyses featuring twenty to thirty word scene context descriptions. These annotations document photographic methodologies, meticulously noting camera angles, for example, eye-level, elevated, low, or bird’s-eye, alongside classifications of photographic style and identifications of textual elements within each visual composition.
    
\begin{figure}[h]
    \centering
    \includegraphics[width=0.75\textwidth]{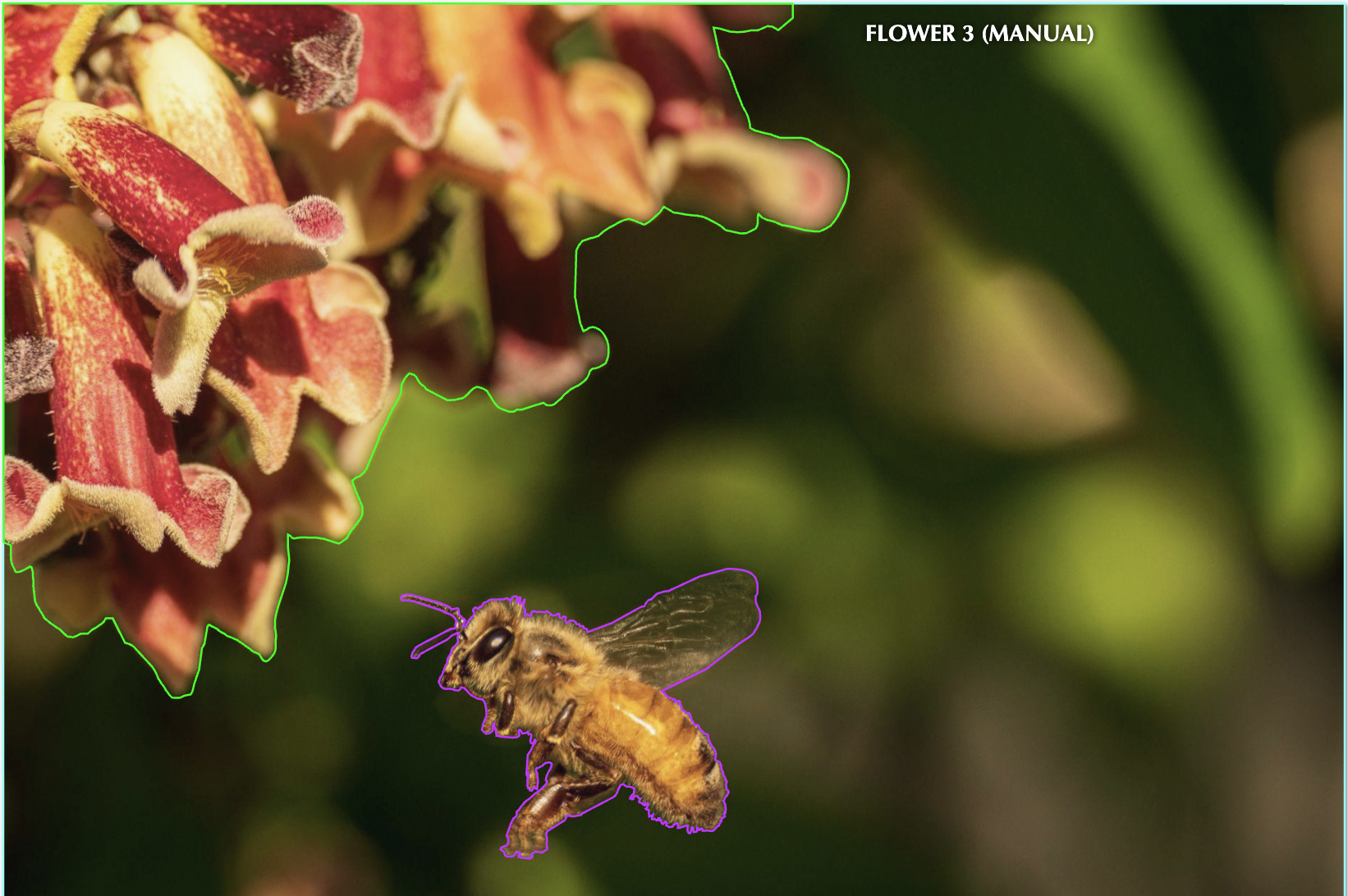}
    \caption[Sample annotated image from DSD]{Sample from our dataset showing an image and its corresponding human segmentation}
    \label{fig:dataset_sample}
    \end{figure}

    
    Figure~\ref{fig:dataset_sample} shows a representative specimen from the DSD corpus. Each datapoint is comprised of a high-resolution image coupled with a structured technical annotation designed to capture salient features of photographic practice.  These annotations systematically document core compositional elements, including lighting dynamics, camera angle, color palette, and the overarching visual arrangement:
    \begin{itemize}
        \item Title: honeybee in the field
        \item Image description: A flying honeybee approaches a group of red and yellow blooms, its delicate wings beating rapidly. The tiny legs of the bee dangle suspended, ready to touch down, as its fuzzy golden body radiates with sunlight.
        \item Scene description: Close-up macro shot, straight-on angle. This perspective allows for an intimate view of the bee’s details and the flower’s textures, highlighting the precision and beauty of the natural world.
    \end{itemize}
    
    \subsubsection{Visual and Semantic Insights from Image Scene Descriptions}
    
    The dataset's annotated captions provide a rich source of semantic and compositional cues that were explored through various visualizations. 
    
\begin{figure}[H]
        \centering
        \includegraphics[width=0.55\linewidth]{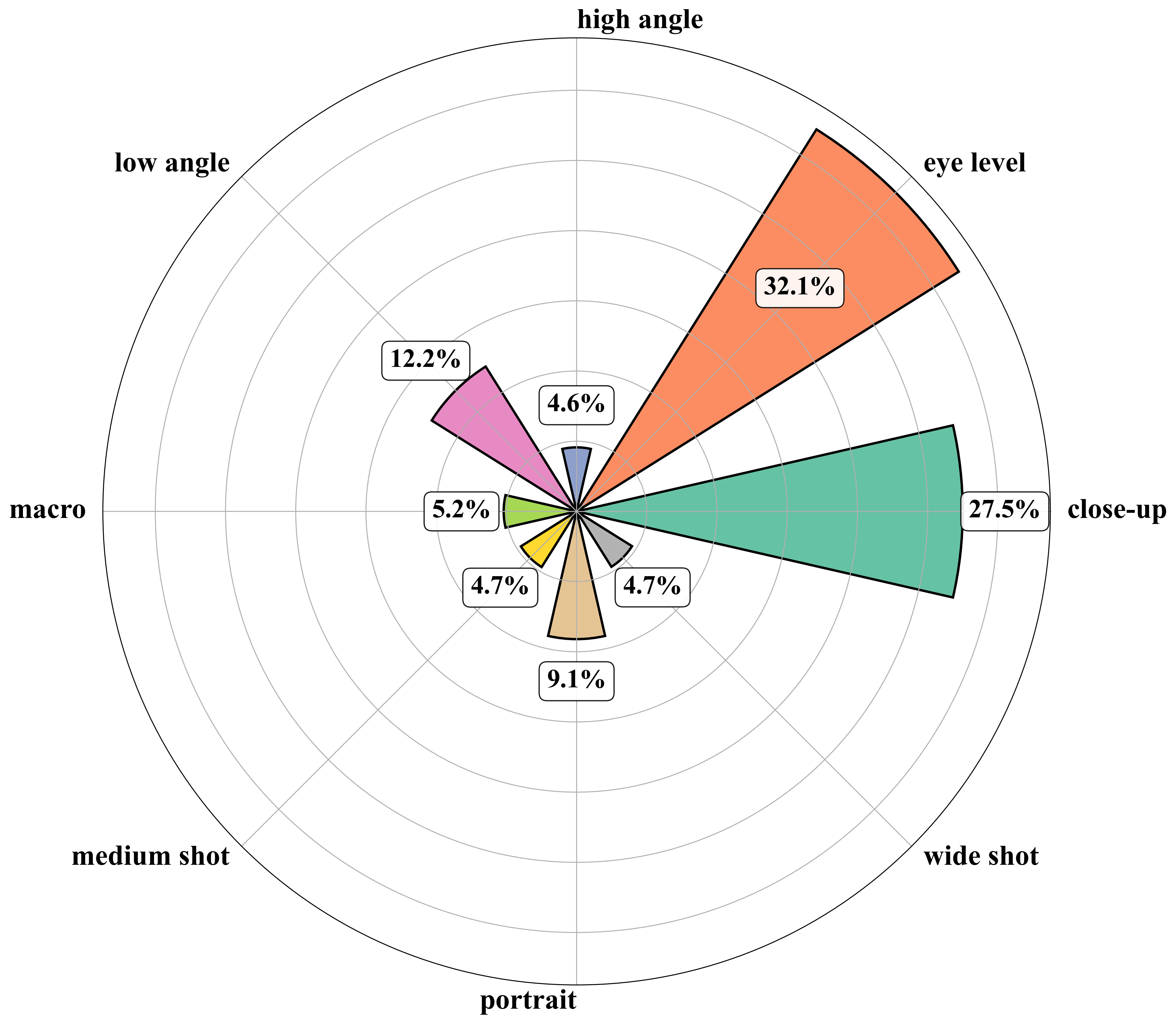}
        \caption[Image shot types]{Shot types radial bar chart showing proportional distribution of different camera angles.}
        \label{fig:shottypes}
    \end{figure}

    Analyzing the 10,610 annotated images from the dataset, as demonstrated in Figure \ref{fig:shottypes}, reveals the distribution of photographic framing styles used in the dataset. Notably, \emph{close-up} and \emph{eye-level} shots dominate the collection, collectively accounting for a significant proportion of the total images. These are followed by \emph{portraits} and \emph{wide shots}, indicating a human preference for human-centric or object-focused compositions. Less common are specialized angles such as \emph{medium shot} and \emph{high angle}, suggesting that extreme perspectives are less frequently captured or annotated.
    
\begin{figure}[H]
        \centering
        \includegraphics[width=0.55\linewidth]{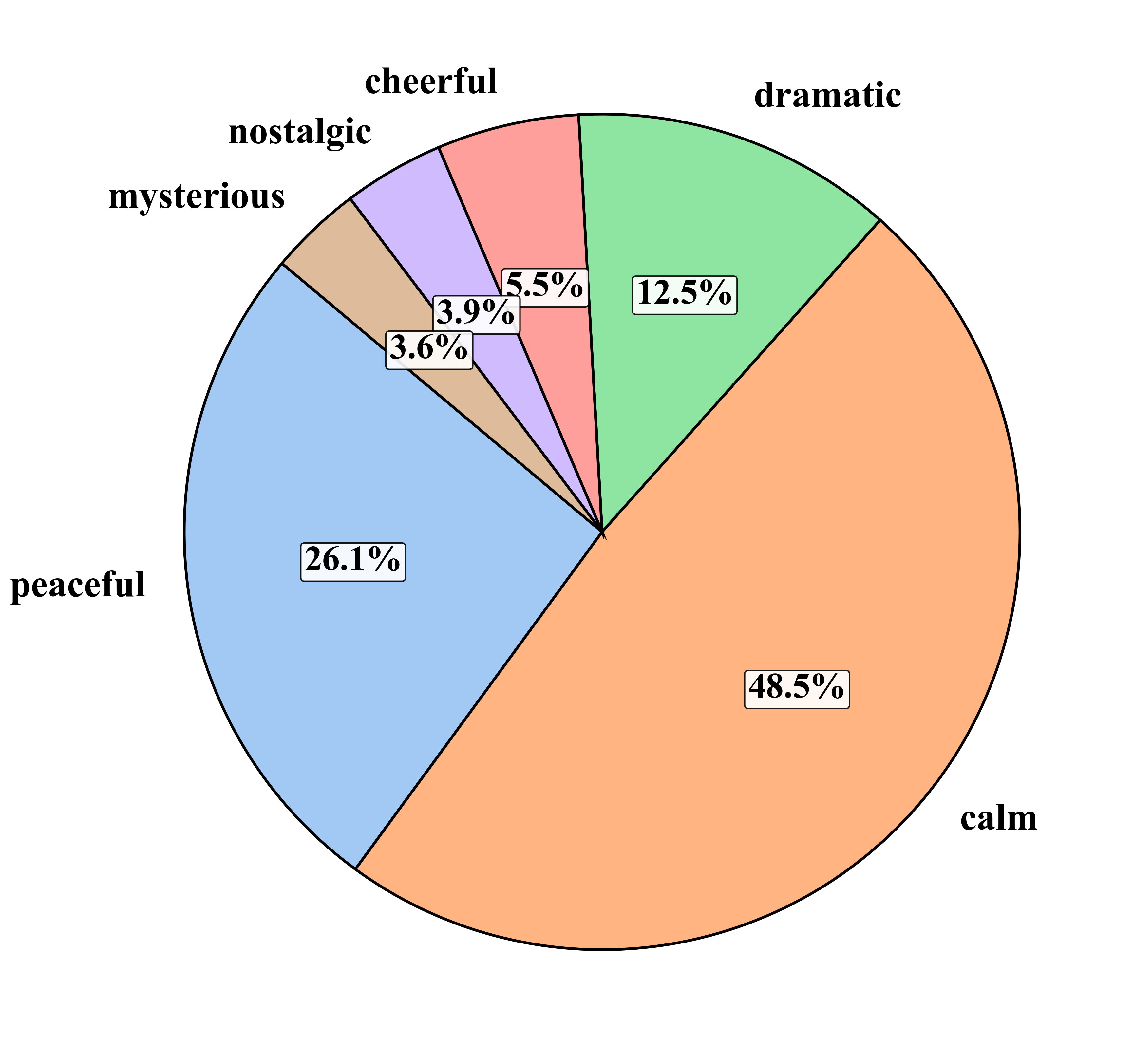}
        \caption[Mood distribution of the images]{Mood distribution pie chart highlighting the relative emotional tone across captions.}
        \label{fig:moodpie}
    \end{figure}

    The mood distribution pie chart (Figure~\ref{fig:moodpie}) provides insight into the emotional tones conveyed by the imagery. \emph{Calm}, \emph{peaceful}, and \emph{dramatic} moods together make up the majority, reflecting an overarching aesthetic of serenity and positivity within the dataset. Other emotional descriptors like \emph{mysterious} and \emph{nostalgic} appear less frequently but contribute to the diversity of affective expression captured through visual storytelling.
    
\begin{figure}[H]
        \centering
        \includegraphics[width=0.9\linewidth]{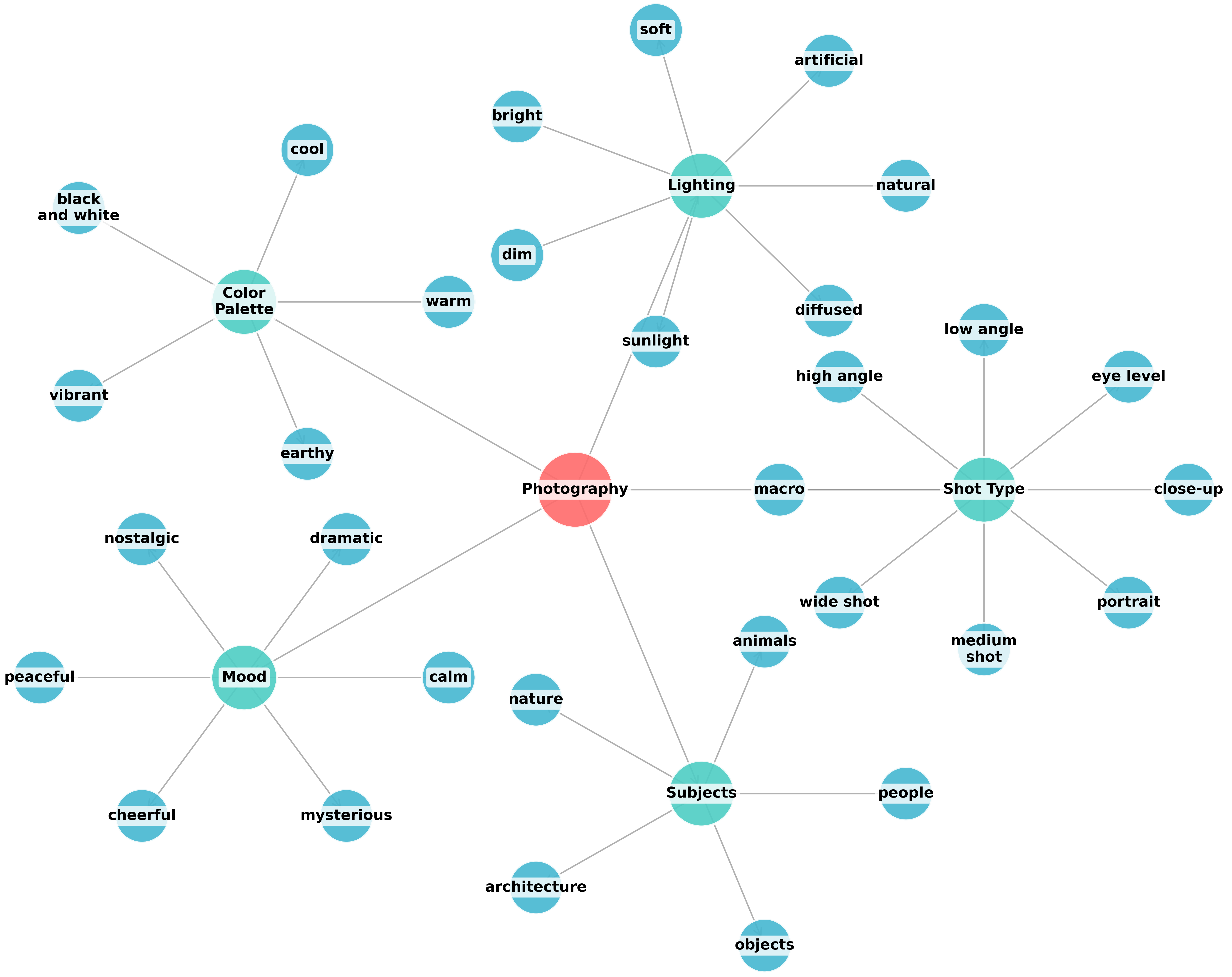}
        \caption[Ontology graph organizing key captions in the images of the dataset]{Ontology graph organizing key caption concepts under high-level photography themes.}
        \label{fig:ontology}
    \end{figure}

    Finally, a structured ontology graph (Figure~\ref{fig:ontology}) was developed to contextualize the various thematic elements identified in the captions. At the root lies the category \emph{Photography} dominates, branching into high-level concepts such as \emph{Shot Type}, \emph{Lighting}, \emph{Color Palette}, \emph{Mood}, and \emph{Subjects}. Each branch further diverges into specialized terms, illustrating the hierarchical and interconnected nature of visual semantics. This ontology not only facilitates a better understanding of the dataset's conceptual landscape but also enables future efforts in categorization, retrieval, and automated annotation.
    
    Together, these visualizations offer a multidimensional understanding of the image dataset, combining technical framing, emotional resonance, and conceptual structure to illuminate patterns that would otherwise remain implicit.
    
    \subsubsection{Label Distribution and Co-occurrence Analysis}
\begin{figure}[h]
        \centering
        \includegraphics[width=0.9\linewidth]{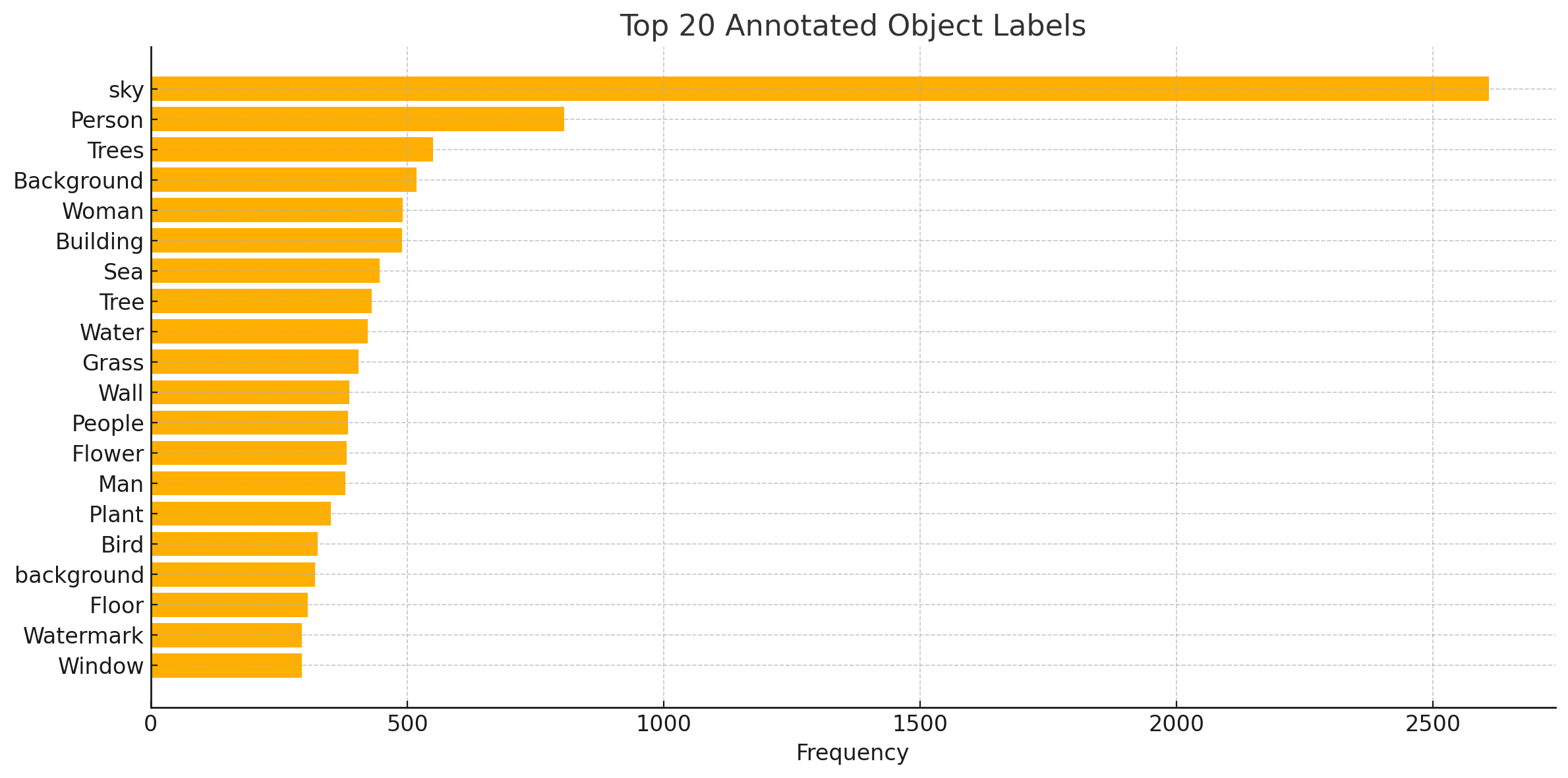}
        \caption[Top 20 most frequently annotated object labels in the DSD dataset.]{Top 20 most frequently annotated object labels in the DSD dataset. The dominance of both human-related and environmental labels indicates a well-rounded annotation scope.}
        \label{fig:label_histogram}
    \end{figure}

    To understand the distribution and semantic relationships within our annotated dataset, we analyzed the frequency and co-occurrence patterns of object labels across images from internal DSD. Figure~\ref{fig:label_histogram} presents the top 20 most frequently annotated object categories, highlighting a dominance of everyday visual elements such as \textit{Person}, \textit{Sky} and \textit{Trees}. The high occurrence of environment-related classes like \textit{Water}, \textit{Grass}, and \textit{See} suggests the dataset includes a rich diversity of outdoor scenes. Similarly, person-centric labels such as \textit{People} and \textit{Man} indicate a substantial presence of human subjects with fine-grained annotations focused on clothing and accessories.
    
\begin{figure}[h]
        \centering
        \includegraphics[width=0.9\linewidth]{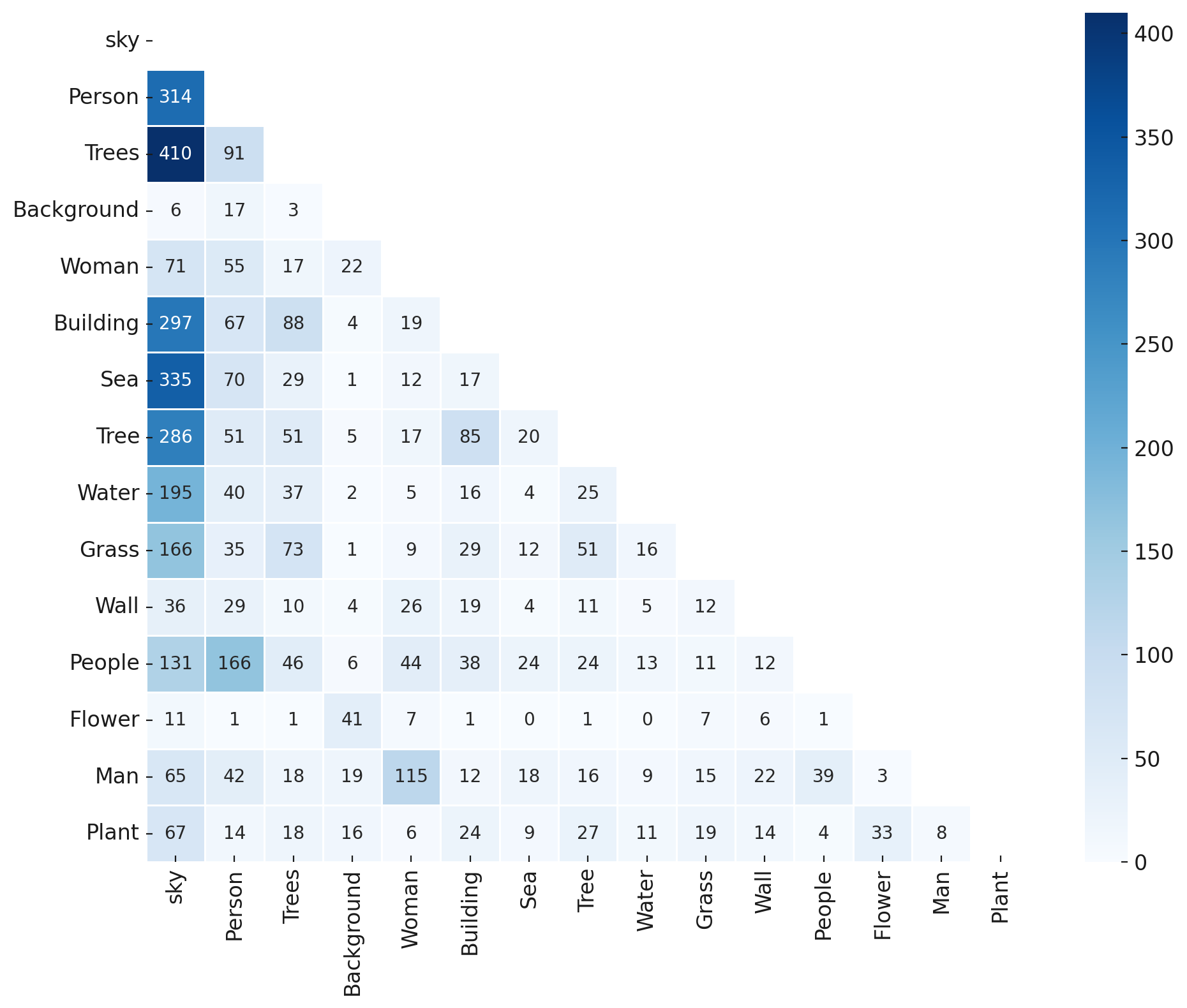}
        \caption[Co-occurrence heatmap of the top 15 labels.]{Co-occurrence heatmap of the top 15 labels. Darker and higher-valued cells indicate stronger co-occurrence across images, suggesting semantic clustering opportunities.}
        \label{fig:cooccurrence_heatmap}
    \end{figure}

    To complement this frequency-based analysis, Figure~\ref{fig:cooccurrence_heatmap} provides a lower triangle co-occurrence heatmap for the top 15 labels. This visualization reveals strong pairwise correlations that reflect common spatial or semantic groupings in natural scenes. For example, \textit{Sky} co-occurs frequently with \textit{Trees} and \textit{Grass}, reflecting structured outdoor settings. Additionally, the co-occurrence pattern between \textit{Man} and \textit{Woman} labels in the dataset shows a significant association with 115 instances where both appear together.  These co-occurrence patterns support the construction of an ontology or label hierarchy, where strongly connected classes may form semantic subgroups. Overall, this dual analysis of label frequency and co-occurrence provides valuable insights into the dataset's coverage, annotation depth, and potential for training structured perception models.
    
\begin{figure}[h]
        \centering
        \includegraphics[width=0.6\linewidth]{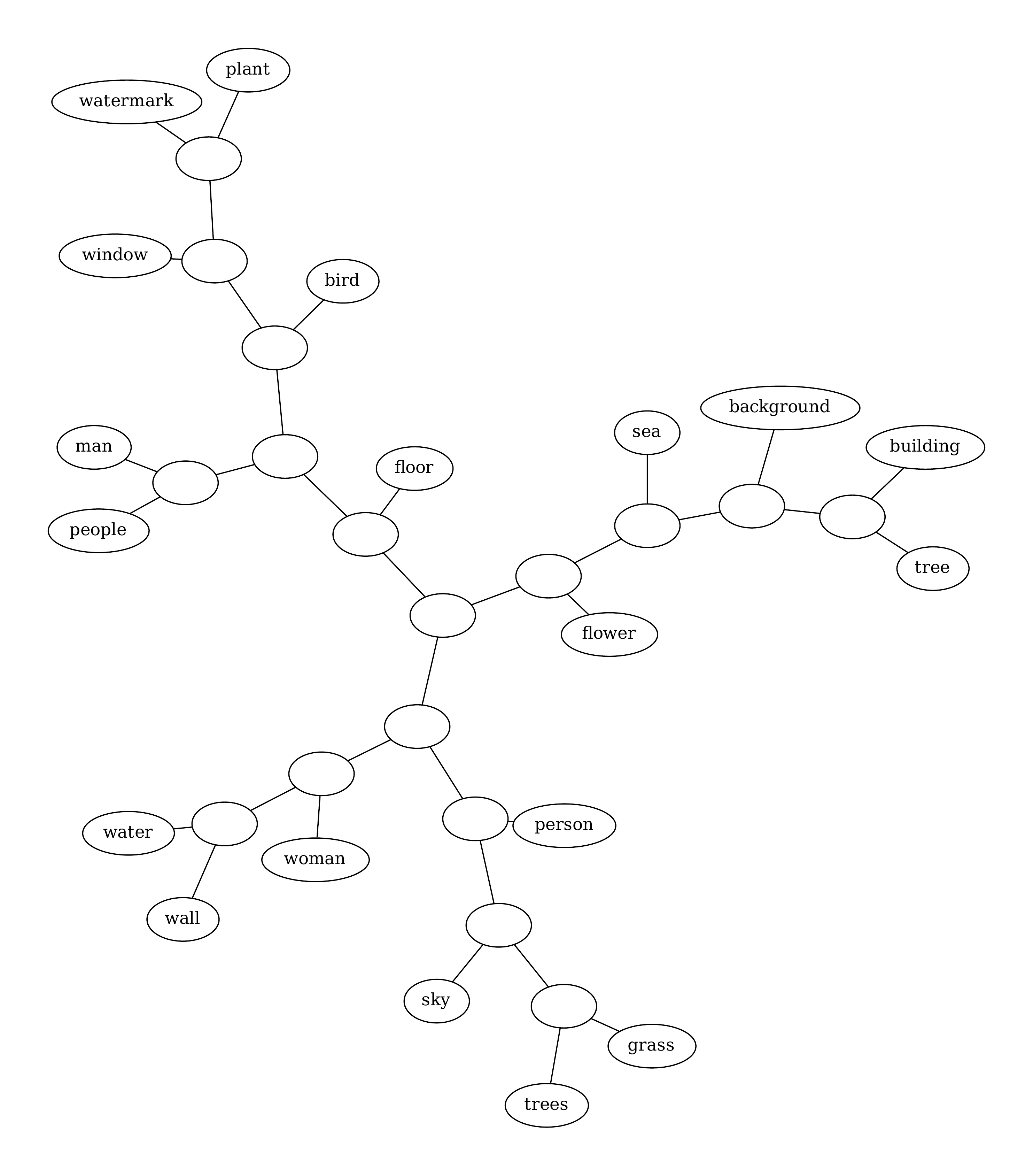}
        \caption[Latent semantic distances between 20 words.]{Latent semantic distances between 20 words in our image scene description.}
        \label{fig:word_graph}
    \end{figure}

    
    Using Normalized Web Distance (NWD) (\cite{vitanyi2010normalized}), we measured the latent semantic distances between the top 20 words in our image scene description dataset. This distance metric evaluates both word pair co-occurrence rates and individual word frequencies in captions using a binary bag-of-words approach. To visualize the resulting distance matrix, we employed the Minimum Quartet Tree Cost (MQTC) algorithm (\cite{cilibrasi2011fast}), a discrete optimization quartet method, to create an optimized un-rooted binary tree (dendrogram) shown in Figure \ref{fig:word_graph}. This tree represents a hierarchical semantic clustering of the 20 search terms, which appear as leaf nodes, based on relationships derived from the complete caption corpus.
    
    \section{Benchmarking the DSD}
    
    \subsection{Experiment Design: Hypotheses and Objectives}
    
    Informed by recent scholarship on data-centric model building, particularly that which was expounded upon by \cite{bhatt2024data}, our experiment methodology was designed to empirically test two core hypotheses. First, we hypothesize that commercial image tagging and scene description systems, though scalable and widely adopted, display significant semantic gaps relative to human-curated annotations, particularly for dense or technically detailed and aesthetically nuanced images. Second, we propose that high-quality, peer-ranked, richly annotated data, such as that which is found in the DSD, can meaningfully improve the performance of state-of-the-art multimodal vision-language models, particularly with respect to scene analysis.
    
    Our overarching objective is to demonstrate that data quality can serve as a first-order lever for improving model outputs.  In particular, our goal is to evaluate the value of the DSD as a high-fidelity dataset and showcase the impact of adopting a data-centric methodology in the context of real-world image analysis tasks. 
    
    In order to operationalize towards this goal, we structured our evaluation along two parallel axes: 
    
    \begin{itemize}
        \item \textbf{Baseline Annotation Evaluation:}
        
        We first assessed the performance of what is arguably the most widely used off-the-shelf commercial system, Amazon's AWS Rekognition label detection API\footnote{\url{https://docs.aws.amazon.com/rekognition/latest/dg/labels-detect-labels-image.html}}, against DSD's human-curated annotations.  This analysis allows us to quantify precision, recall, and label alignment gaps, thereby identifying areas where even a state-of-the art generalist system like Rekognition underperforms when compared to human-evaluated data. 
        
        \item \textbf{Fine-Tuned Model Evaluation:}
    
        Next we fine-tuned two multi-modal foundational models LLaVVA-NeXT (\cite{llava_next_blog}) and BLIP (\cite{li2023blip}), two leading open-source multimodal vision-language models, using DSD's structured data. This enabled us to measure not only direct performance improvements using known benchmarks but also to assess the degree to which the high-quality, human-in-the-loop annotated data contained as part of the DSD could accelerate model specialization for technical scene analysis tasks. For our comparative analysis, we deliberately selected LLaVA-NEXT and BLIP2 as representative models from different architectural paradigms in vision-language modeling. LLaVA-NEXT, with its modular design that combines a vision encoder with a large language model (LLM) backbone, represents the newer generation of vision-language models that leverage the capabilities of powerful pretrained LLMs through lightweight adapter layers. This architecture maintains strong alignment between visual features and language representations throughout the model. BLIP2, on the other hand, employs a Q-Former architecture that serves as an intermediary between the vision encoder and language model, creating a bottleneck that helps bridge the modality gap. By examining these architecturally distinct approaches, we aimed to investigate how different model designs respond to fine-tuning on technical image descriptions. This comparison provides valuable insights into the robustness and adaptability of different architectural choices when optimizing for domain-specific image captioning tasks.
    \end{itemize}
    
    Through this multi-step experiment design, we aim not only to benchmark performance improvements but also to surface deeper insights into the interaction between data quality, annotation type, and model architecture. Ultimately, our results seek to validate the premise that data is not just a supporting resource, but the principal substrate for enabling precision AI. Of course, this evaluation strategy is far from the only one available.  The diversity and depth of the various elements of the DSD lend themselves to a plethora of unaddressed evaluation opportunities, which we actively support and encourage.
    
    \subsection{Dataset Description and Preparation}
    For our experiments, we randomly sampled 10,610 images from the  DataSeeds.AI catalog. These images included the full range of photographic styles, subjects, and technical properties represented in the larger DataSeeds.AI dataset. Each image is accompanied by its complete annotation package, including concise descriptive titles, detailed narrative descriptions (15+ words), technical scene analyses (20-30 words), and semantic segmentation masks.
    
    The data was partitioned into a 90/10 split between training and validation sets. Each datapoint was formatted as an input-output pair, structured to simulate the type of conversational prompting we assessed would be technically useful and also the expected response from the model, which was taken directly from the DSD scene description: 
    
    \begin{enumerate}
        \item Split the dataset into training (90\%) and validation (10\%) sets
        \item Format each example as an input-output pair:
        \begin{itemize}
            \item Input: Image + prompt requesting scene analysis
            \item Output: Human-annotated scene analysis from our dataset
        \end{itemize}
    \end{enumerate}
    
    Our training set consists of 9,549 image-text pairs, while the validation set contains 1,061 pairs. We ensured that there was no overlap between sets to prevent data contamination. Each sample of the formatted dataset is organized in a conversation-based format where there is a conversation based on the user's prompt requesting technical scene analysis, and also the expected response from the model, which is taken directly from the annotated database scene description. For example, each entry from the dataset for the LLAVA-NEXT model is organized as follows:
    \newpage
    \begin{verbatim}
      {
        "image": "guru_10k_img/0ccfc9a8-1547-4077-
        8fc1-f04e77d21043.jpg",
        "conversations": [
          {
            "from": "human",
            "value": "<image>\nDescribe this scene
            (at least 20 to 30 words and not more than 80 words):
            focus on the overall context, environment,
            lighting, camera angle (eye level/high/low/bird's eye),
            color palette, photography style,
            and any text visible in the image."
          },
          {
            "from": "gpt",
            "value": "Close-up macro shot, straight-on angle.
            This perspective allows for an intimate
            view of the bee's details and the
            flower's textures, highlighting
            the precision and beauty of the natural world."
          }
        ]
      },
    \end{verbatim}
    
    \subsection{Fine-Tuning Methodology}
    
    \subsubsection{LLAVA-NEXT Model Selection and Training Configuration}
    
    We selected LLAVA-NEXT as our base model due to its strong performance on vision-language tasks and efficient fine-tuning capabilities. Specifically, we utilized the smaller language model variant with 0.5 billion parameters \footnote{lmms-lab/llava-onevision-qwen2-0.5b-ov} (\cite{llava_onevision}) along with the Siglip \footnote{google/siglip-so400m-patch14-384}  vision encoder (\cite{zhai2023sigmoid}). This combination offers several advantages for our scene analysis task:
    
    \begin{itemize}
        \item Enhanced visual representations through the SigLIP vision encoder's hierarchical feature extraction
        \item Improved text generation quality despite the smaller language model size
        \item Support for detailed technical analysis through its pre-training regime
        \item Efficient parameter-tuning methodologies requiring reasonable computational resources
    \end{itemize}
    
    Following best practices for efficient adaptation \cite{peft} we adopted a Parameter-Efficient Fine-Tuning (PEFT) approach, specifically employing LoRA (\cite{hu2022lora}) with the following settings: 
    
    \begin{enumerate}
        \item Initialize the model with pretrained lmms-lab/llava-onevision-qwen2-0.5b-ov weights
        \item Apply LoRA to reduce trainable parameters:
        \begin{itemize}
            \item LoRA rank: 32
            \item LoRA alpha: 32
            \item LoRA dropout: 0.1
        \end{itemize}
        \item Focus tuning on specific components for fine tuning: adapter and the language model
    \end{enumerate}
    
    Table~\ref{tab:training_config} details the hyperparameters and settings used for our fine-tuning experiment. We employed mixed precision training using bfloat16 format to optimize memory usage and employed gradient checkpointing for additional efficiency. The training was conducted on a single NVIDIA A100 40GB GPU with a batch size of 2, compensated by 8 gradient accumulation steps to effectively simulate larger batches. The base model architecture consists of approximately 916.97 MB of parameters in total. During fine-tuning, only about 513.47 MB of parameters are trained, representing approximately 56\% of the total model parameters, while the remaining parameters remain frozen. The model uses BF16 precision. The multimodal projector adapter contains 1,836,800 parameters that are 100\% trainable in our configuration, consisting of two linear layers with a GELU (Gaussian Error Linear Units) activation in between. We used the LoRA technique with rank 32 and alpha 32, training for 3 epochs with a learning rate of 1e-5 and cosine learning rate scheduler.
    
    \begin{table}[h]
    \centering
    \caption{Hyperparameters and Configuration for Fine-tuning LLAVA-NEXT}
    \label{tab:training_config}
    \begin{tabular}{@{}ll@{}}
    \hline
    \textbf{Parameter} & \textbf{Value} \\
    \hline
    Learning rate & 1e-5 \\
    Optimizer & AdamW \\
    Learning rate schedule & Cosine decay \\
    Warmup ratio & 0.03 \\
    Weight decay & 0.01 \\
    Batch size & 2 \\
    Gradient accumulation steps & 8 \\
    Training epochs & 3 \\
    Maximum sequence length & 8192 \\
    Maximum gradient norm & 0.5 \\
    Precision & BFloat16 \\
    \hline
    \end{tabular}
    \end{table}
    
    The model is trained with a consistent system prompt template across all samples, requesting the scene description of the image from the model. During training, we carefully monitored model performance by evaluating on the validation set every 50 steps, rather than only at epoch boundaries. This frequent evaluation strategy allowed us to precisely identify the optimal checkpoint, capturing performance peaks that might otherwise be missed with less frequent validation. We configured the training to automatically save the model with the lowest validation loss, ensuring that our final model represented the best generalization performance rather than simply the final training state.
    
    \subsubsection{BLIP2 Experimental Protocol and Training Configuration}
    For fine-tuning BLIP2 (OPT 2.7B variant architecture), we used the `caption\_coco\_opt2.7b` pre-trained weights. We fine-tuned the model for the captioning task using our dataset. Gradient checkpointing was enabled, but the Vision Transformer (ViT) component was not frozen (`freeze\_vit: False`), allowing its parameters to be updated during training. The objective was to adapt the model to generate technical scene descriptions matching our annotations.
    
    The BLIP2 fine-tuning was configured with an initial learning rate of 1e-5 using the AdamW optimizer and a linear warmup followed by cosine decay schedule (17 warmup steps). The training ran for 10 epochs with a batch size of 8 per single Nvidia A100 80GB device and gradient accumulation of 1. Weight decay was set to 0.01. Mixed precision (AMP) was enabled. The target generation length was set between 8 and 100 tokens, using beam search with 5 beams. Specific image and text processors (`blip2\_image\_train`, `blip\_image\_eval`, `blip\_caption`) were used for data handling with an image size of 364x364.
    
    Unlike LLaVA-NeXT, where the best checkpoint was selected based on minimum validation loss, the BLIP2 checkpoint selection relied on an aggregate metric evaluated on the validation set. This aggregate metric (`val\_agg\_metrics=CIDEr + Bleu\_4`) is calculated based on a combination of evaluation scores. We monitored this metric across epochs and found that it peaked at epoch 1 with a value of 0.2626. Therefore, the checkpoint from epoch 1 was selected as the final fine-tuned BLIP2 model for evaluation. The training loss showed a consistent decrease throughout the 10 epochs, starting at 2.780 in the first epoch and ending at 1.692 in the final epoch.
    
    \subsection{Evaluation Setup}
    
    \subsubsection{AWS Rekognition Evaluation}
    As part of an initial review of dataset quality, we evaluated AWS' Rekognition label detection API against the DSD. While Rekognition provided a scalable solution during the early operational stages of the GuruShots platform, its suitability for dataset curation at training-grade standards (i.e. the diversity of data demands by model builders) necessitated an empirical reassessment. 
    
    Using a confidence threshold of 50\%, we conducted a comparative analysis between Rekognition outputs and human-annotated ground truth across a sample of 10,610 images. Our evaluation metrics included:
    
    \begin{itemize}
        \item Precision, recall, and F1 scores between Rekognition outputs and human annotations
        \item Analysis of label overlap statistics between machine and human annotations
        \item Examination of the relationship between image complexity (number of objects) and annotation quality
    \end{itemize}
    
    This evaluation allowed us to quantify the semantic gap between automated commercial tagging systems and human perception, providing a clear baseline against which to measure the value of human-in-the-loop annotation.
    
    \subsubsection{Multi-modal Model Evaluation Metrics}
    We employed the following metrics to evaluate the generated descriptions from the fine-tuned model against human annotations. The BLEU (\cite{papineni2002bleu}) score measures the precision of n-gram matches between generated and reference texts, with particular emphasis on 4-gram matches (BLEU-4) to capture phrase-level accuracy. The ROUGE-L (\cite{lin2004rouge}) metric evaluates the recall of the longest common subsequences, providing insight into the structural similarity between generated and reference descriptions. Additionally, we utilized BERTScore (\cite{zhang2019bertscore}), which leverages contextual embeddings from pre-trained language models to compute semantic similarity beyond surface-level token matching.
    
    For comparing generated descriptions with ground truth annotations, we calculate:
    
    \begin{equation}
        \text{BLEU-4}(G, R) = BP \cdot \exp\left(\sum_{n=1}^{4} w_n \log p_n\right)
    \end{equation}
    
    where $G$ represents the generated text, $R$ the reference annotation, $p_n$ is the n-gram precision (proportion of matching n-grams), $w_n$ are weights for each n-gram precision (typically uniform weights), and $BP$ is the brevity penalty that penalizes overly short generations:
    
    \begin{equation}
        BP = 
        \begin{cases}
            1 & \text{if } |G| \geq |R| \\
            e^{(1-|R|/|G|)} & \text{if } |G| < |R|
        \end{cases}
    \end{equation}
    
    For ROUGE-L, we calculate:
    
    \begin{equation}
        \text{ROUGE-L} = \frac{LCS(G, R)}{\text{length}(R)}
    \end{equation}
    
    where $LCS(G, R)$ is the length of the longest common subsequence between the generated text $G$ and reference $R$. Unlike n-gram matching, $LCS$ allows for non-contiguous matches while preserving sequence order, making it particularly suitable for evaluating fluency and coherence.
    
    For BERTScore, we compute the F1 score between contextual embeddings of the generated and reference texts:
    
    \begin{equation}
        \text{BERTScore}_{\text{F1}}(G, R) = 2 \cdot \frac{\text{P}_{\text{BERT}} \cdot \text{R}_{\text{BERT}}}{\text{P}_{\text{BERT}} + \text{R}_{\text{BERT}}}
    \end{equation}
    
    where $\text{P}_{\text{BERT}}$ and $\text{R}_{\text{BERT}}$ are the precision and recall scores computed using token-level cosine similarities between BERT embeddings:
    
    \begin{equation}
        \text{P}_{\text{BERT}} = \frac{1}{|G|} \sum_{g_i \in G} \max_{r_j \in R} \cos(e(g_i), e(r_j))
    \end{equation}
    
    \begin{equation}
        \text{R}_{\text{BERT}} = \frac{1}{|R|} \sum_{r_j \in R} \max_{g_i \in G} \cos(e(g_i), e(r_j))
    \end{equation}
    
    Here, $e(\cdot)$ represents the contextual embedding of a token, and $\cos(\cdot,\cdot)$ is the cosine similarity. BERTScore captures semantic similarity and accounts for synonyms, paraphrases, and related concepts that may be missed by lexical-matching metrics like BLEU and ROUGE, making it particularly valuable for evaluating descriptions where different terminology might convey the same technical information.
    
    We also leveraged the enhanced capabilities of Long-CLIP (\cite{zhang2024long}) to compute accurate semantic alignment scores between images and their textual descriptions, including longer and more detailed captions to assess the alignment of images, and also the descriptions generated by the LLAVA-NEXT and BLIP2 base and fine-tuned models. The CLIPscore metric was calculated by measuring the cosine similarity between normalized image and text embeddings, scaled to a percentage value. Specifically, given an image path and a text description, we first loaded our fine-tuned Long-CLIP model, which supports extended token sequences beyond the original CLIP's 77-token limitation. We used this model since the contributions generated by the model in our experiments might be more than 77 tokens. For each evaluation, we preprocess the input image $I$ through the model's transformation pipeline and tokenize the text description $T$. The model then produces embedding vectors $\mathbf{v_I} = E_I(I)$ and $\mathbf{v_T} = E_T(T)$ through its respective encoders. These embeddings are $L_2$-normalized to unit vectors: $\hat{\mathbf{v_I}} = \frac{\mathbf{v_I}}{||\mathbf{v_I}||}$ and $\hat{\mathbf{v_T}} = \frac{\mathbf{v_T}}{||\mathbf{v_T}||}$. The final CLIPscore is computed as: $\text{CLIPscore} = 100 \times \cos(\hat{\mathbf{v_I}}, \hat{\mathbf{v_T}})$, where higher values indicate stronger semantic alignment between the image content and textual description, benefiting particularly from Long-CLIP's ability to process and understand fine-grained attributes and relationships in detailed captions.
    
    \subsection{Results and Analysis}
    
    \subsubsection{Performance Analysis of AWS Rekognition Label Detection}
    Using a confidence threshold of 50\%, we conducted a comparative analysis between Rekognition outputs and human-annotated ground truth across a sample of 10,610 images. 
    Table~\ref{tab:overall_performance} summarizes the key performance metrics.
    
    \begin{table}[h]
    \centering
    \caption{Overall Performance Metrics of AWS Rekognition Label Detection}
    \label{tab:overall_performance}
    \begin{tabular}{lr}
    \hline
    \textbf{Metric} & \textbf{Value} \\
    \hline
    Total Images Analyzed & 10,610 \\
    Average Precision & 0.1359 \\
    Average Recall & 0.4232 \\
    Average F1 Score & 0.1916 \\
    Unique AI Labels & 2,335 \\
    Unique Human Labels & 7,926 \\
    Label Overlap & 1,503 (18.96\% of human labels) \\
    \hline
    \end{tabular}
    \end{table}
    
\begin{figure}[htbp]
        \centering
        \begin{subfigure}[b]{0.49\textwidth}
            \centering
            \includegraphics[width=\textwidth]{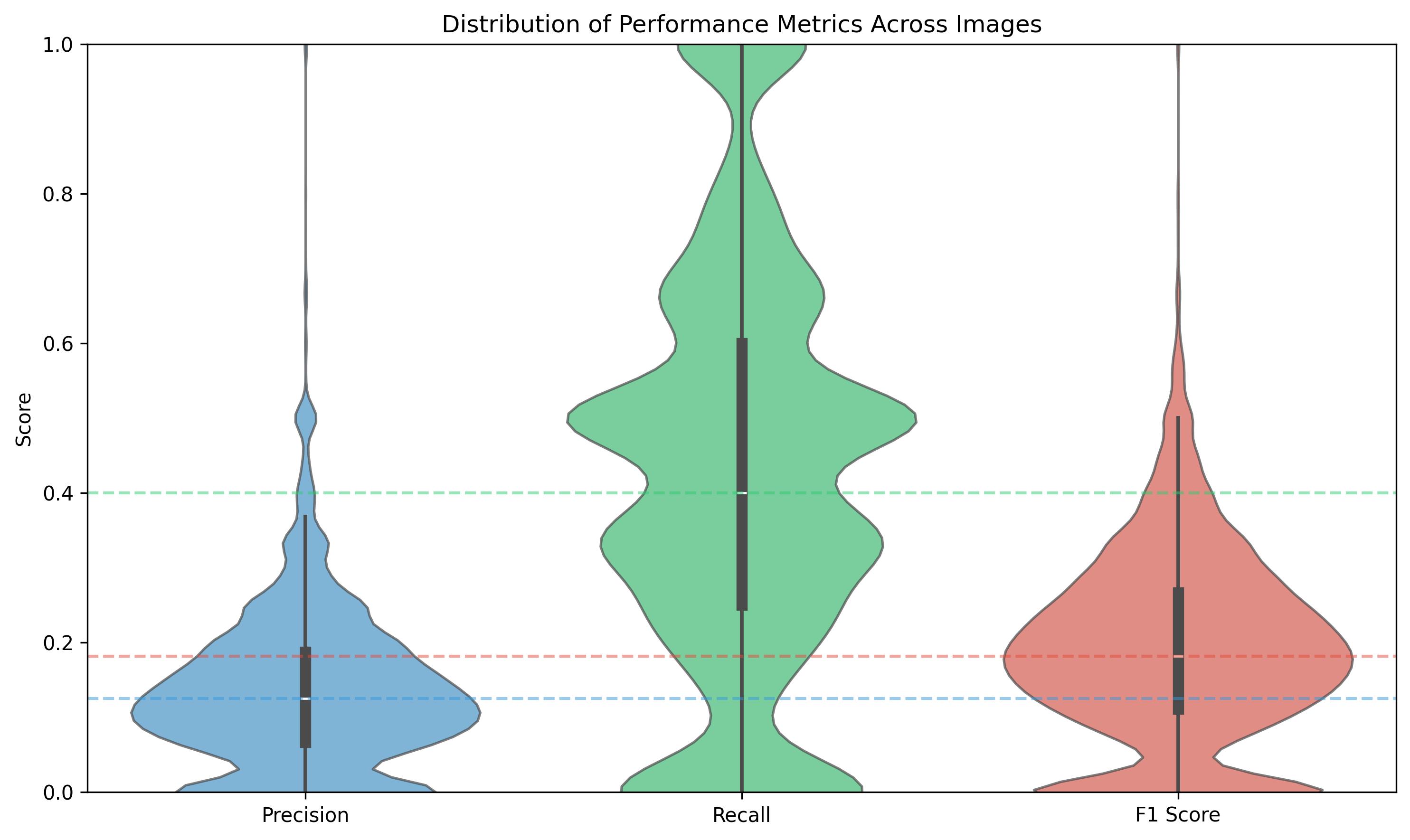}
            \caption{Distribution of precision, recall, and F1 scores across all analyzed images.}
            \label{fig:metric_distribution}
        \end{subfigure}
        \hfill
        \begin{subfigure}[b]{0.49\textwidth}
            \centering
            \includegraphics[width=\textwidth]{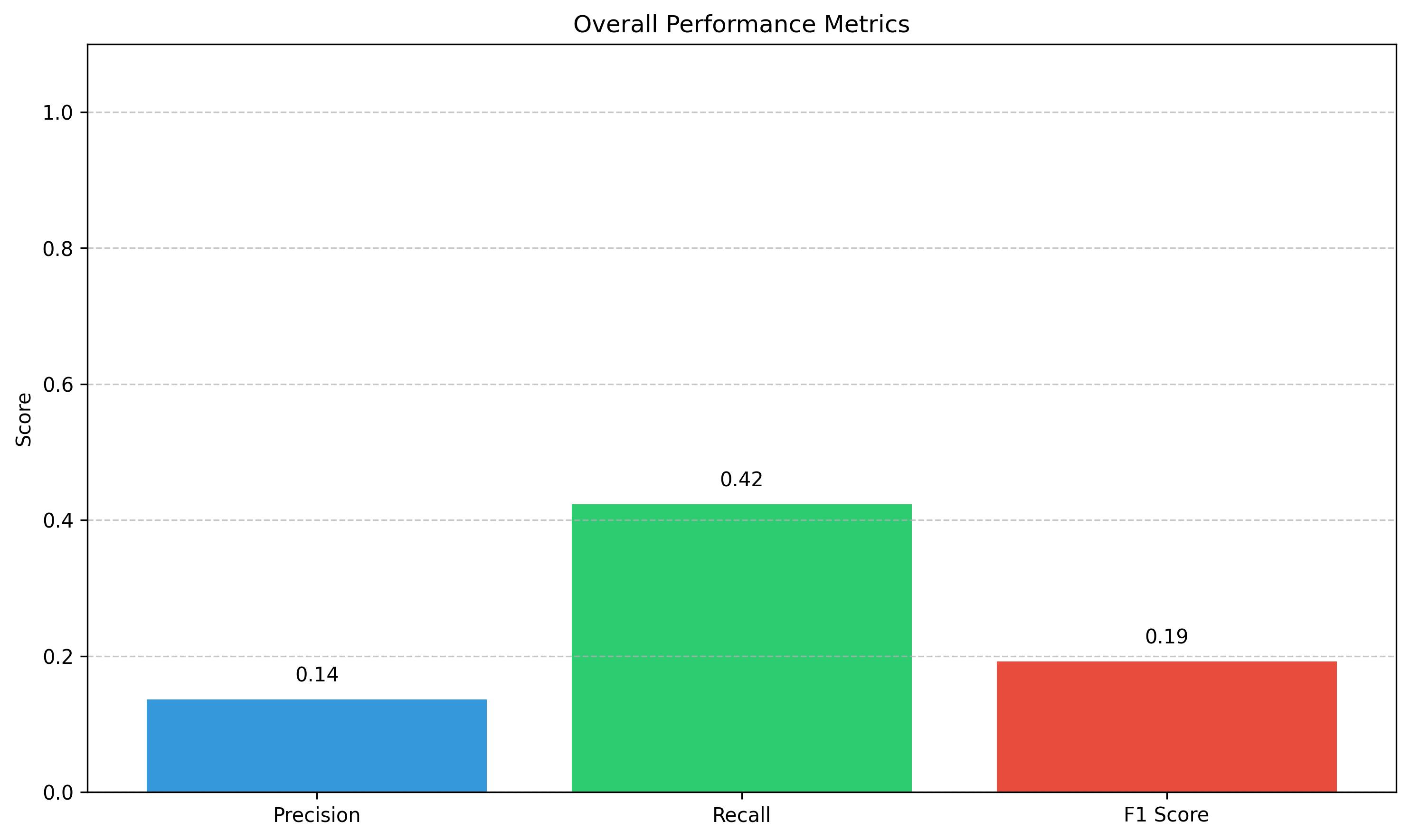}
            \caption{Summary of average precision, recall, and F1 score across the dataset.}
            \label{fig:overall_metrics}
        \end{subfigure}
        
        \parbox{0.8\textwidth}{
            \centering
            \caption{Analysis of precision, recall, and F1 scores for the dataset evaluation.}
            \label{fig:combined_metrics}
        }
    \end{figure}

    The evaluation revealed a consistent pattern. While Rekognition demonstrated reasonable recall (0.4232), indicating moderate sensitivity to human-identified features, precision, which is arguably the most important component, remained substantially lower (0.1359), leading to a higher incidence of false positives. This imbalance is clearly illustrated in the distribution analysis set forth in Figure~\ref{fig:metric_distribution}, which shows a wide distribution for recall compared to precision, with some images achieving recall values approaching 1.0.
    
\begin{figure}
    \centering
    \includegraphics[width=0.75\textwidth]{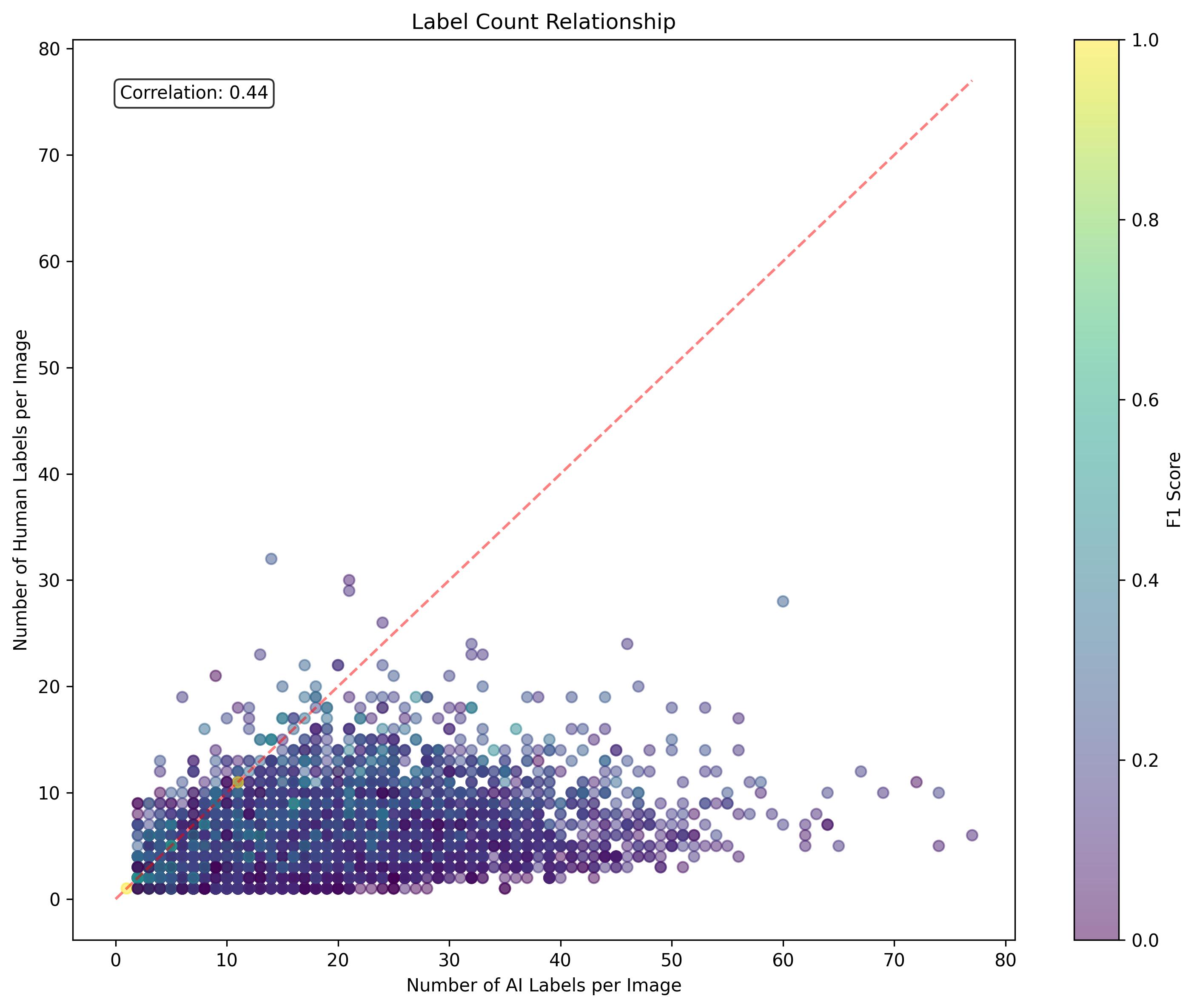}
    \caption[Relationship between AI and human label counts per image]{Relationship between AI and human label counts per image, with point colors indicating F1 scores.}
    \label{fig:label_diversity}
    \end{figure}

    Further inspection of the label count scatterplot (Figure~\ref{fig:label_diversity}) and the headline metrics tells a clearer story when we weigh precision and recall together. Rekognition's outputs do not scale with image complexity (r=0.44), and it often produces 10–30 tags where humans supply fewer than 10. Yet what matters for downstream utility is not the sheer number of tags but their joint correctness captured by the F1 score. Despite the extra labels, Rekognition's overall F1 is only 0.19. In the plot, the densest clusters of points sit in the low F1 region, and even the most label-rich images rarely break above an F1 of $\approx$0.3. This indicates that the recall gains from overgeneration are more than canceled out by the accompanying precision losses. In short, breadth without balance adds noise faster than it adds true positives, so researchers should pay closer attention to F1 (and downstream task impact) rather than label count alone when assessing model value.
    
    The aggregate F1 score of 0.19 (Figure~\ref{fig:overall_metrics}) reflects the harmonic mean of these divergent precision-recall curves, further highlighting the limitation of applying Rekognition outputs natively for supervised learning tasks. The performance pattern is consistent with the Rekognition's confidence threshold setting (50\%), which was deliberately selected to maximize detection coverage rather than precision. Of the 7,926 unique human labels, only 1,503 (18.96\%) were recovered through machine inference, underscoring a substantial semantic gap between automated outputs and human perception. 
    
    These findings corroborate prior, often unheeded, warnings that commercially available tagging systems often struggle with domain-specific datasets that require fine-tuned semantic understanding. Our results suggest that high-quality human-in-the-loop annotation remains critical for dataset construction when the target objective is a robust turn-key dataset, particularly for tasks that require nuanced aesthetic, technical, or compositional judgements.
    
    \subsubsection{LLAVA-NEXT Training Dynamics and Results}
    
\begin{figure}[h]
    \centering
    \includegraphics[width=0.9\textwidth]{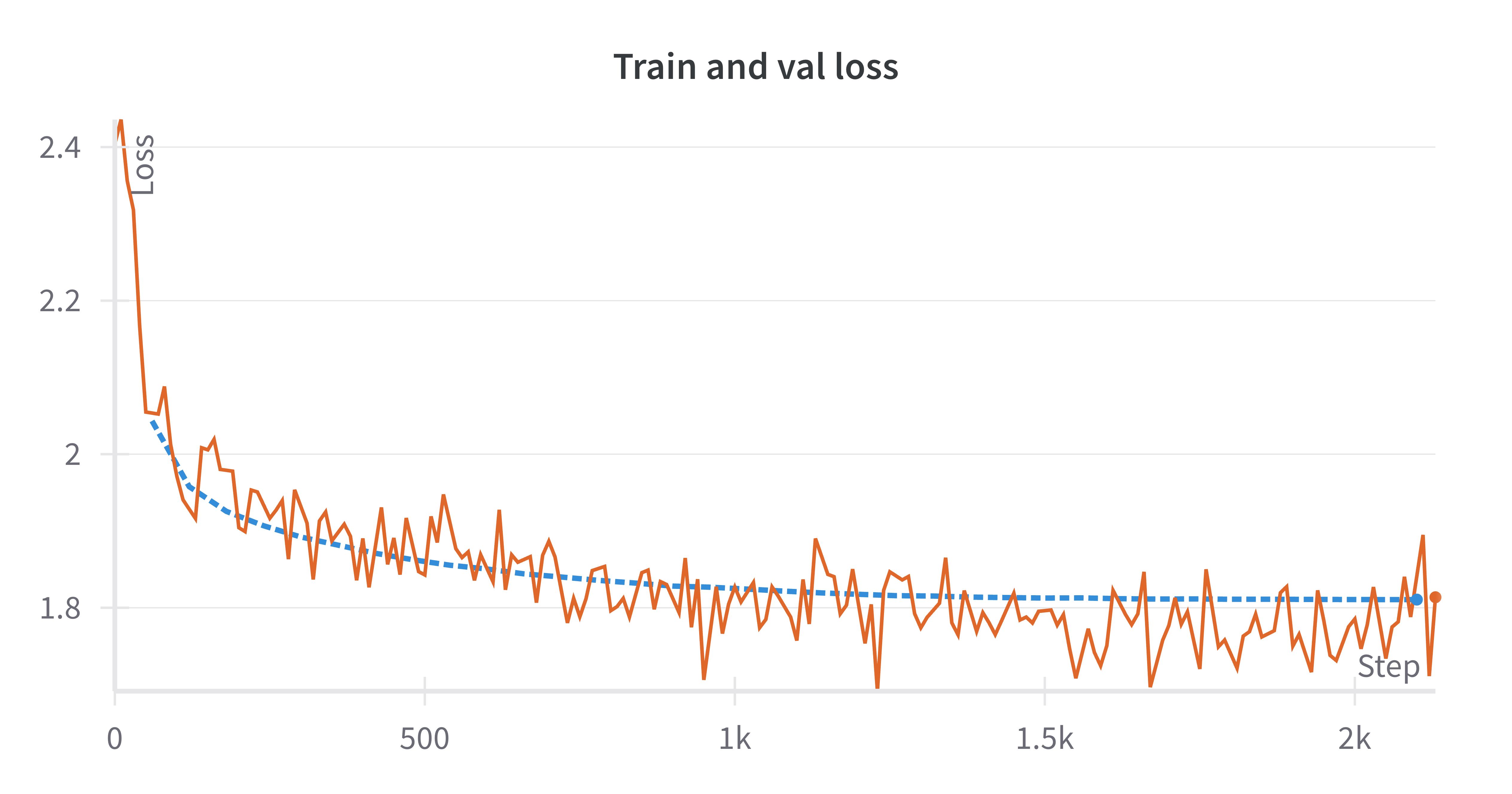}
    \caption[Training and validation loss curves during fine-tuning of LLAVA-NEXT]{Training and validation loss curves during fine-tuning of LLAVA-NEXT are shown, with training loss represented by an orange solid line and validation loss by a blue dashed line. The model reaches optimal performance with a validation loss of 1.83 at step 1750 (epoch 2.9), after which continued training yields minimal improvement despite further decrease in training loss.}
    \label{fig:loss_curves}
    \end{figure}

    Figure~\ref{fig:loss_curves} illustrates the training and validation loss curves throughout the LLaVA-NeXT fine-tuning process. The training loss (orange solid line) shows a steady decrease from the initial value of approximately 2.4 to around 1.7 by the end of training, indicating continuous improvement in the model's ability to fit the training data. The validation loss (blue dashed line) also follows a similar pattern in the course of training. 
    
    Based on the validation loss, we selected the checkpoint from step 1,750 as our final LLaVA-NeXT model for evaluation, as it demonstrates the best generalization performance on unseen data. Our fine-tuning process was completed in 30 hours. At inference time, the fine-tuned model processed images at 2.30 seconds per sample using the same GPU.
    
    Table~\ref{tab:finetune_results} demonstrates the substantial benefits of fine-tuning LLAVA-NEXT, with significant improvements across all evaluation metrics. Most notably, we observed a remarkable 24.09\% relative improvement in BLEU-4 scores, indicating a dramatic enhancement in n-gram precision in the generated descriptions. This substantial improvement suggests that our fine-tuned model produces considerably more accurate word sequences that better match human references.
    
    \begin{table}[h]
    \centering
    \caption{Performance Comparison: Base vs. Fine-tuned LLAVA-NEXT}
    \label{tab:finetune_results}
    \begin{tabular}{@{}lcccc@{}}
    \hline
    \textbf{Model} & \textbf{BLEU-4} & \textbf{ROUGE-L} & \textbf{BERTScore} & \textbf{CLIPScore} \\
    \hline
    LLAVA-NEXT (Base) & 0.0199 & 0.2089 & 0.2751 & 0.3247 \\
    LLAVA-NEXT (Fine-tuned) & 0.0246 & 0.2140 & 0.2789 & 0.3260 \\
    \hline
    Absolute Improvement & 0.0048 & 0.0051 & 0.0039 & 0.0013 \\
    Relative Improvement (\%) & 24.09 & 2.44 & 1.40 & 0.41 \\
    \hline
    \end{tabular}
    \end{table}
    
    The ROUGE-L metric shows a meaningful 2.44\% improvement, indicating enhanced capture of sequential information and longer matching sequences between generated text and reference descriptions. This improvement in ROUGE-L, combined with the BLEU-4 gains, demonstrates that fine-tuning significantly improved the model's ability to generate descriptions with better linguistic structure and sequence.
    
    The BERTScore metric shows a solid 1.40\% improvement, indicating that fine-tuning has enhanced the model's semantic understanding capabilities. This improvement is particularly significant as BERTScore captures deeper contextual relationships beyond surface-level word matching. Additionally, the CLIPScore shows a positive change (0.41\%), suggesting that the model maintains and slightly improves its visual-semantic alignment while gaining these substantial textual improvements.
    
    The consistent pattern of improvements across all metrics demonstrates that our fine-tuning approach effectively enhances the model's descriptive capabilities without compromising its fundamental visual understanding. The particularly strong improvement in BLEU-4 scores indicates that our fine-tuning method is especially effective at improving precision in image descriptions, a critical aspect for applications requiring accurate visual information communication. These results validate our fine-tuning approach and highlight its effectiveness in optimizing LLAVA-NEXT for more precise, semantically rich, and structurally coherent image descriptions.
    
    \subsubsection{BLIP2 Fine-Tuning Results}
    
    The results for BLIP2 fine-tuning are presented in Table~\ref{tab:blip2_finetune_results}. Similar to LLaVA-NeXT, BLIP2 shows significant improvements in lexical overlap metrics after fine-tuning, with BLEU-4 increasing from 0.001 to 0.047 and ROUGE-L improving from 0.126 to 0.242. The extremely high relative improvement for BLEU-4 (4600\%) is primarily due to the very low baseline score (0.001), which makes even modest absolute improvements appear dramatic in percentage terms. This mathematical artifact occurs when calculating relative improvements from near-zero baselines. Nevertheless, the absolute improvement of 0.046 in BLEU-4 and 0.116 in ROUGE-L indicates that fine-tuning effectively adapted BLIP2 to generate text that better matches the n-gram patterns and subsequences present in the reference descriptions. 
    
    \begin{table}[h]
    \centering
    \caption{Performance Comparison: Base vs. Fine-tuned BLIP2 (OPT 2.7B)}
    \label{tab:blip2_finetune_results}
    \begin{tabular}{@{}lcccccc@{}}
    \hline
    \textbf{Model} & \textbf{BLEU-4} & \textbf{ROUGE-L} & \textbf{BERTScore} & \textbf{CLIPScore} \\
    \hline
    BLIP2 (Base) & 0.001 & 0.126 & 0.0545 & 0.2854 \\
    BLIP2 (Fine-tuned) & 0.047 & 0.242 & -0.0537 & 0.2583 \\
    \hline
    Absolute Improvement & 0.046 & 0.116 & -0.1082 & -0.0271 \\
    Relative Improvement (\%) & 4600.00 & 92.06 & -198.53 & -9.49 \\
    \hline
    \end{tabular}
    \end{table}
    
    However, unlike LLaVA-NeXT, the BERTScore drastically decreased from 0.0545 to -0.0537 (-198.53\% relative change). It’s important to understand that BERTScore evaluates semantic similarity using cosine similarities between contextual embeddings of tokens in the generated and reference texts, rather than relying on exact lexical matches like traditional classification F1 scores (which are inherently non-negative). In this case, the BERTScore is suggesting a significant divergence in semantic meaning or phrasing style between the fine-tuned BLIP2 outputs and the reference annotations, despite better lexical overlap. The negative BERTScore indicates performance below the baseline expectation for semantic similarity. The CLIPScore also showed a notable decrease from 0.2854 to 0.2583 (-9.49\% relative change), which is more substantial than previously observed, indicating a moderate reduction in overall image-text semantic alignment after fine-tuning.
    
    \subsubsection{Comparative Analysis of Model Architectures}
    
    When comparing the fine-tuning effects on both models evaluated on the same validation set, we observe distinct performance patterns. LLAVA-NEXT demonstrates consistent improvements across all metrics, with modest but balanced gains: 24.09\% in BLEU-4, 2.44\% in ROUGE-L, 1.40\% in BERTScore, and 0.41\% in CLIPScore. In contrast, BLIP2 shows a more dramatic but uneven response to fine-tuning. While BLIP2 achieves substantially larger improvements in lexical metrics (4600\% in BLEU-4 and 92.06\% in ROUGE-L), it simultaneously experiences significant degradation in semantic understanding (-198.53\% in BERTScore) and image-text alignment (-9.49\% in CLIPScore). The extreme BLEU-4 improvement for BLIP2 stems primarily from its near-zero baseline (0.001), making the relative improvement percentage less meaningful than the absolute gain of 0.046. 
    
    These contrasting patterns suggest that LLAVA-NEXT's architecture better preserves semantic understanding and visual grounding during fine-tuning, whereas BLIP2's fine-tuning process appears to optimize for lexical matching at the expense of semantic coherence and visual alignment. This trade-off highlights the importance of considering multiple evaluation dimensions when assessing multimodal model performance, as improvements in surface-level text similarity do not necessarily correlate with enhanced semantic understanding or visual-linguistic alignment.
    
    \subsubsection{Qualitative Analysis}
    To illustrate the improvements achieved through fine-tuning of LLAVA-NEXT and BLIP2 models, we present a comparison between descriptions generated by the base and fine-tuned models for the same image. Figure~\ref{fig:model_comparison} shows an example where there are notable differences in the generated descriptions.
    
    \begin{figure}[h]
    \centering
    \includegraphics[width=0.75\textwidth]{files/sony_headphones.jpg}
    \caption{Comparison of base and fine-tuned LLAVA-NEXT and BLIP2 model outputs. The image shows a pair of Sony WH-1000XM3 headphones with black finish and copper accents.}
    \label{fig:model_comparison}
    \end{figure}
    
    \begin{table}[h]
    \centering
    \caption{Comparison of Generated Descriptions from Base and Fine-tuned Models}
    \label{tab:generation_comparison}
    \begin{tabular}{@{}p{14cm}@{}}
    \hline
    \rowcolor[gray]{0.9}
    \textbf{Reference Description} \\
    It's a close-up shot of a sleek pair of Sony WH-1000XM3 headphones resting on a dark surface. The color palette features matte black, deep gray, and subtle copper accents. The lighting is soft and directional, creating gentle highlights and shadows that emphasize the headphones' contours and premium texture. The mood is sleek, modern, and sophisticated, evoking a sense of quiet focus and premium quality. \\
    \hline
    \rowcolor[gray]{0.9}
    \textbf{LLAVA-NEXT Model} \\
    \hline
    \textbf{Base Model:} The image showcases a pair of sleek, black Sony headphones with orange accents. The headphones are resting on their sides against a dark background, highlighting the brand name ``Sony'' and model number ``\textbf{WH-1003}''. A close-up shot captures the ear cups in detail, showcasing ergonomic design and padding for comfort during extended use. \\
    \hline
    \textbf{Fine-tuned Model:} The image shows a pair of black Sony headphones with orange accents, resting on a dark surface. The camera angle is low and slightly above the headphones to provide an eye-level perspective. The lighting appears soft and diffused, likely from ambient light sources in a room or studio setting. There are no visible texts other than ``\textbf{WH-100XM3}'' printed near the ear cup area. \\
    \hline
    \rowcolor[gray]{0.9}
    \textbf{BLIP2 Model} \\
    \hline
    \textbf{Base Model:} A pair of headphones sitting on top of each other. \\
    \hline
    \textbf{Fine-tuned Model:} A pair of headphones taken from a close-up angle, highlighting the details of the headphones. The lighting is soft, highlighting the details of the headphones. \\
    \hline
    \end{tabular}
    \end{table}
    
    The improvements introduced through our fine-tuning experiments reveal significant enhancements across multiple dimensions, as evidenced in Table~\ref{tab:generation_comparison} and supported by our quantitative metrics. Our fine-tuned model achieved substantial improvements across all evaluation metrics, with a particularly noteworthy 24.09\% relative improvement in BLEU-4 scores, indicating significantly enhanced n-gram precision. The qualitative comparison of the Sony headphones image demonstrates several key improvements:
    
    First, the fine-tuned model exhibits superior product identification accuracy by correctly identifying the model number (``WH-1000XM3'') compared to the base model's incorrect identification (``WH-1003''). This precise product recognition represents a critical advancement for applications requiring accurate catalog information or product descriptions.
    
    Second, the fine-tuned model provides more accurate compositional details, describing both the positioning of the headphones and the camera perspective with greater fidelity to the reference. The description of lighting as ``soft and diffused'' more closely aligns with the reference's ``soft and directional'' characterization, demonstrating enhanced perception of visual ambiance.
    
    Third, the fine-tuned model demonstrates greater contextual awareness by describing environmental elements like the surface supporting the headphones and the overall photographic setup, rather than focusing exclusively on product features like ``ergonomic design and padding'' mentioned by the base model.
    
    These improvements align with our quantitative findings across all metrics. Beyond the substantial BLEU-4 improvement, we observed modest but consistent gains in ROUGE-L (2.44\%), BERTScore (1.40\%), and CLIPScore (0.41\%). The pattern of improvements suggests that our fine-tuning approach enhances the model's capacity to generate descriptions that are not only more factually accurate but also better aligned with human-like perceptual attention to compositional elements, lighting conditions, and contextual details.
    
    While both models still diverge from capturing the full aesthetic qualities mentioned in the reference (such as the ``sleek, modern, and sophisticated mood''), the fine-tuned model's enhanced accuracy on technical details and environmental context represents a meaningful step toward more reliable and precise visual-linguistic understanding.
    
    The BLIP2 fine-tuning results present an interesting case that helps explain our quantitative metrics. Despite the concerning drop in BERTScore from 0.0545 to -0.0537 and CLIPScore from 0.2854 to 0.2583, the qualitative example reveals a nuanced picture of how lexical improvements can coexist with semantic shifts.
    
    As shown in Table~\ref{tab:generation_comparison}, the base BLIP2 model produces a minimal description, identifying only the basic object category and their relative positioning (``sitting on top of each other''). The description lacks any details about the photographic composition, lighting, or specific product characteristics.
    
    The fine-tuned BLIP2 model demonstrates notable expansion in descriptive elements, particularly concerning photographic aspects. It correctly identifies the ``close-up angle'' and comments twice on the ``soft lighting'' that highlights details. This repeated emphasis on lighting and photographic perspective aligns with n-gram patterns in the reference description (which mentions ``soft and directional'' lighting and ``close-up shot''), explaining the substantial improvements in BLEU-4 (4600\% relative increase) and ROUGE-L (92.06\% relative increase).
    
    However, this increased attention to photographic elements comes with two significant trade-offs. First, the fine-tuned model fails to identify specific product information (Sony WH-1000\-XM3), the copper accents, or the premium quality aspects present in the reference. Second, and more notably, there is a marked repetition in the description ("highlighting the details of the headphones" appears twice), suggesting a degradation in the model's semantic coherence and natural language generation capabilities.
    
    This repetitive pattern and focus shift help explain the dramatic decrease in BERTScore (-198.53\%), as the model has moved away from balanced, semantically cohesive descriptions toward a more formulaic approach that emphasizes certain aspects while neglecting others. The decrease in CLIPScore (-9.49\%) further suggests that while the model has learned to include more photographic terminology, it has simultaneously lost some ability to capture the overall semantic relationship between the image content and generated text.
    
    This example illustrates how BLIP2's fine-tuning process has resulted in a model that has effectively learned n-gram patterns associated with photographic description but at the cost of semantic coherence and balanced content representation—a trade-off not observed in the more stable LLAVA-NEXT architecture.
    
    \section{Conclusion}
    Taken together, the results presented in this study provide compelling evidence that carefully curated, peer-ranked, and richly annotated datasets like the DSD can significantly enhance the performance and reliability of modern vision-language models. The fine-tuning experiments conducted with LLAVA-NEXT and BLIP2 demonstrate that the DSD not only improves syntactic precision and semantic fidelity but also enhances the overall alignment between visual content and textual descriptions. These improvements are critical for advancing AI models that require deep contextual understanding, precise scene interpretation, and human-like perceptual reasoning.
    Importantly, the success of the DSD’s multi-tier human-in-the-loop approach, which integrates human perceptual judgments with technical scene descriptions, validates the importance of AI researchers shifting towards a data-centric approach to AI model development.  Further, this approach challenges conventional reliance on noisy, crowd-sourced datasets by offering a higher-fidelity alternative that captures nuanced visual semantics and aesthetic preferences.  In doing so, the DSD aligns closely with emerging trends in data-centric AI development, where the quality, diversity, and contextual richness of training data are increasingly being recognized as primary drivers of model performance.      
    
    The broader implications of this work extend far beyond the immediate improvements demonstrated in this study. The DSD, representing only a small fraction of the 100 million-plus images available through DataSeeds.AI, has the potential to serve as a foundational resource for a wide range of AI applications, including multimodal retrieval, technical scene analysis, and generative image understanding. As the AI community continues to grapple with the challenges of building models that can reason about complex, real-world visual scenes, datasets like the DSD will play an essential role in pushing the boundaries of what these models can achieve.
    
    Uniquely, the commercial infrastructure underpinning the DSD, the DataSeeds.AI platform, presents a critical opportunity for data-centric model builders. Beyond its catalog, DataSeeds.AI is positioned to fulfill bespoke image data requests through its expansive global network of semi-professional photographers. This on-demand capability, where datasets can be curated to precise specifications and delivered with full licensing rights and annotation, marks a significant evolution in how training data can be provisioned for AI development. DataSeeds.AI introduces a flexible, responsive pipeline for fine-tuning that directly addresses the domain-specific edge-case needs of contemporary multimodal models.
    
    The DSD's architecture illustrates a best-in-class example of a data-centric model in action. Our results underscore the transformative power of aligning data collection processes with the operational realities and representational demands of modern AI systems. Whether in support of autonomous vision tasks, stylized content generation, or nuanced scene comprehension and object identification, the DSD and its underlying infrastructure demonstrate that a combination of high-quality static data and responsive dynamic sourcing can serve as the backbone of scalable, reliable, and performant AI. 
    
    We encourage the broader research community to not only leverage the DSD in their own work but also to explore the rich potential of the broader DataSeeds.AI dataset ecosystem. The future of AI will be shaped not just by the models we build, but by the precision and intentionality of the data we use to train them. We hope that the DSD will not just be another dataset, but rather, a launchpad for a new era of human-aligned intelligence.

    \section*{Acknowledgment}
    The first author gratefully acknowledges Rudi L. Cilibrasi for his valuable insights into computing latent semantic distance, which significantly contributed to the analysis of the image scene description database.

\clearpage
\bibliography{files/references}
\bibliographystyle{plainnat}

\end{document}